\documentclass[journal]{IEEEtran}

\usepackage{amsmath,amssymb}
\usepackage{graphicx}
\usepackage{array}
\usepackage{xcolor}
\usepackage{url}
\newcommand{\toprule}{\hline}
\newcommand{\midrule}{\hline}
\newcommand{\bottomrule}{\hline}

\newcommand{\balance}{}
\usepackage{placeins}

\graphicspath{{figures/}}
\newcommand{\dtcl}{D_{\mathrm{TCL}}}
\newcommand{\drgb}{D_{\mathrm{RGB}}}
\newcommand{\lmap}{\mathcal{L}_{\mathrm{map}}}
\newcommand{\ldice}{\mathcal{L}_{\mathrm{Dice}}}
\newcommand{\lsem}{\mathcal{L}_{\mathrm{sem}}}
\newcommand{\method}{\textsc{SAAF}}

\makeatletter
\renewcommand{\subsection}{%
  \@startsection{subsection}{2}{\z@}%
  {2.0ex plus 0.5ex minus 0.3ex}
  {0.5ex plus 0.2ex}
  {\normalfont\normalsize\itshape}}
\makeatother
\begin{document}
\raggedbottom

\title{From Change Captions to Change Detection: Semantic-Appearance Agreement Framework for Remote Sensing Change Detection}

\author{Yuan~Qian, Jie~Ma%
\thanks{This work was supported in part by the National
Natural Science Foundation of China under Grant 62101052; and in part by the Fundamental Research Funds for the Central Universities under Grant 2024JJ040.     (Corresponding author: Jie Ma.)}%
\thanks{The authors are with the School of Information Science and Technology, Beijing Foreign Studies University, Beijing 100089, China (e-mail: majie\_sist@bfsu.edu.cn).}%
}

\maketitle

\begin{abstract}
Remote sensing change detection (RSCD) is essential for monitoring land-cover changes and urban development. However, most methods demand pixel-level change masks, which are costly and time-consuming to annotate. Weakly supervised methods reduce this cost by using image-level change labels. Yet these labels indicate only whether a change occurs, leaving models to recover the location of the change and semantic meaning through additional and complex mechanisms. This missing information can be supplied directly by change captions, which describe what changes, what it becomes, and where it occurs. Therefore, we introduce change-caption-guided RSCD, using change captions as the sole task-specific supervision to learn change masks without manually annotated change masks. Our framework has two components: a caption-driven generation pipeline that produces bi-temporal remote sensing image pairs at scale with controlled changes matching each caption, and a change detector guided by the caption’s transition semantics. The detector uses our Semantic-Appearance Agreement Framework (SAAF) to combine caption-grounded semantic responses with RGB differences for change localization, while text conditioning guides dense prediction. Experiments on our newly constructed Flair-RSGen dataset and WHU-CDC show that SAAF outperforms the closest reproduced limited-supervision baselines in macro-averaged IoU and F1 under the evaluated protocols. Code is publicly available at \url{https://github.com/qianyuancs/SAAF}.

\end{abstract}


\section{Introduction}
\IEEEPARstart{R}{emote}-sensing change detection (RSCD) identifies
land-surface changes from co-registered images acquired over the
same geographical area at different times. By providing spatially
explicit information and temporally comparable information on surface evolution, RSCD supports urban
growth and land-use planning \cite{lee2021urban}, environmental
monitoring \cite{song2014environment}, and rapid post-disaster
building-damage assessment \cite{zheng2021damage}, thereby facilitating timely and informed decision-making.

Fully convolutional Siamese networks learn paired representations for change prediction~\cite{daudt2018fully}. BIT models spatial--temporal context through compact visual tokens~\cite{chen2022bit}, while ChangeFormer uses a hierarchical Transformer in a Siamese architecture~\cite{bandara2022changeformer}. Despite these advances, training accurate supervised RSCD models commonly requires manually delineated pixel-level change masks.
Constructing such masks is labor-intensive because annotators must
compare two temporal observations, distinguish genuine land-cover
transitions from seasonal or radiometric variations, and delineate
the boundaries of changed regions. This substantial annotation cost has motivated increasing interest in weakly supervised RSCD, in which predictions are learned from less expensive supervisory signals \cite{wslcd,acwcd,transwcd}. 

Most image-level weakly supervised change detection methods follow a
classification-to-localization pipeline. A classifier is first trained
with paired images and binary scene-level labels, after which its
activation maps are extracted and refined into pixel-level change
predictions or pseudo labels. Along this direction, WSLCD combines
image-level supervision with self-supervised representation learning
\cite{wslcd}; ACWCD exploits Transformer self-attention and change
priors to refine CAM-derived pseudo labels \cite{acwcd}; TransWCD
integrates multiscale activation maps with a scene-adaptive predictor
\cite{transwcd}; and SemSAM-CD converts activation evidence into
prompts for segmentation-model refinement \cite{semsamcd}. These
methods have substantially reduced the dependence of RSCD on change-mask annotations for training.
\begin{figure}[!t]
    \centering
    \includegraphics[width=0.96\columnwidth]
    {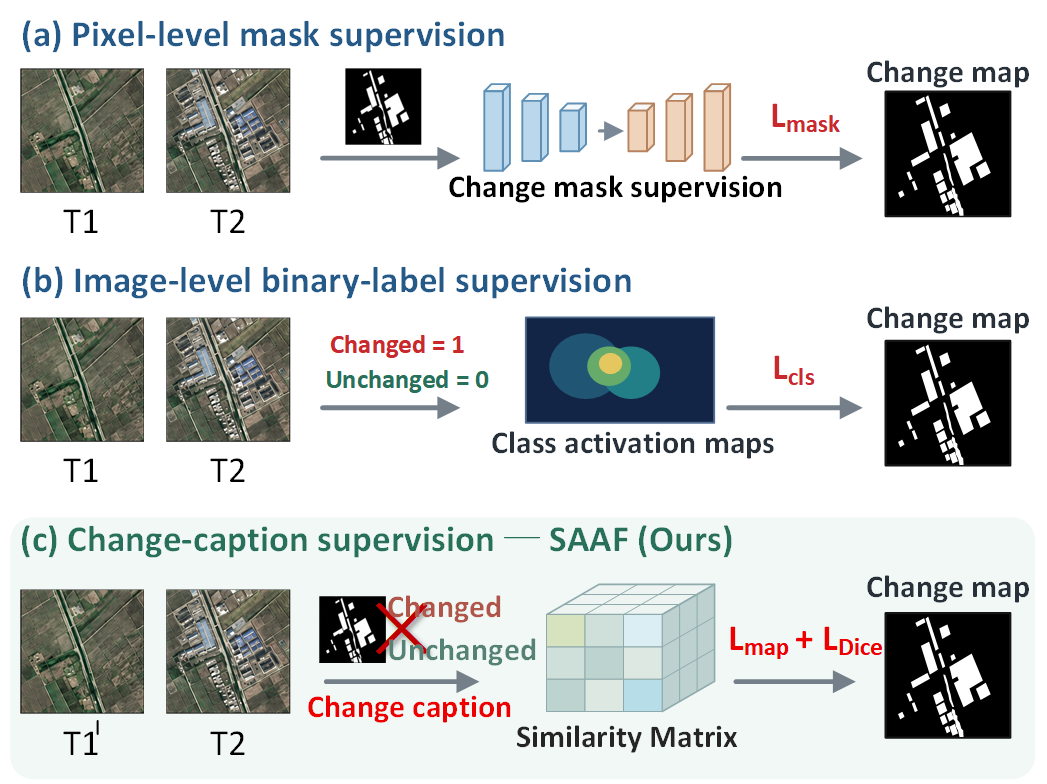}
    \caption{Comparison of supervision for remote sensing change detection:
(a) pixel-level change masks, (b) binary image-level labels,
and (c) change captions used by SAAF.}
    \label{fig:supervision_comparison}
\end{figure}
Nevertheless, binary scene-level supervision only indicates whether
a change exists in an image pair. It does not explicitly describe
which land-cover category disappears, which category emerges, or the
semantic direction of the transition. Consequently, existing methods
must infer both the spatial location and the semantic identity of a
change from image appearance and internal activations alone. They then require more complex models. Their
localization responses also can be distracted by seasonal
variation, illumination differences, phenological changes, or
residual image misregistration, even when no semantically relevant
transition has occurred.

Change captions encode transition semantics unavailable in a binary
changed/unchanged label. For example, the caption ``farmland changes
to built-up land'' identifies the disappearing source class and the
emerging target class, thereby specifying the direction of change.
Recent vision--language RSCD methods incorporate textual cues through
category prompts, caption features, or referring expressions to
improve change representation and localization
\cite{changeclip,mdsnet,referringcd}. However, they are generally
developed under fully or semi-supervised settings and still use
pixel-level change masks to optimize the change predictor. In
contrast, change-captioning methods map bi-temporal images to textual
descriptions \cite{rsicc_dbt,change_captioning}, rather than captions
to change maps.

In this work, we investigate a new RSCD setting in which a
sample-specific change caption serves as the sole supervisory
annotation for training, without change-mask supervision. Compared
with binary image-level labels, change captions provide
source-to-target semantics while avoiding the costly collection of
pixel-level change masks. The key challenge is to transform this
spatially weak semantic description into reliable pixel-level
evidence: A change caption specifies the transition and may
provide coarse spatial cues, but it does not delineate
the changed pixels.

To address this challenge, we propose the Semantic-Appearance
Agreement Framework (\method{}). It employs a frozen,
remote-sensing-adapted Text-grounded Contrastive Learning model
(TCL-DA), built upon CLIP, as a pseudo-mask teacher to associate the
source-to-target semantics of a change caption with the bi-temporal
images \cite{tcl}. The resulting semantic responses are combined with
temporal appearance differences to generate caption-consistent pseudo
change masks. These masks supervise a bi-temporal detector, while
the same caption further conditions its feature learning. In this
way, \method{} converts spatially weak change captions into
supervision by connecting \emph{what} transition is described with
\emph{where} observable change occurs.

In addition, we develop a change-caption-driven method for generating
bi-temporal remote sensing image pairs and use it to construct
Flair-RSGen. Given a change caption, the generation method interprets
its source category, target category, and spatial cue to synthesize
a corresponding second-temporal image, enabling controllable
land-cover transitions.
Extensive experiments are conducted on Flair-RSGen and WHU-CDC
\cite{whucdc}. Quantitative and qualitative results demonstrate that
\method{} outperforms the reproduced baselines in
terms of macro-averaged IoU and F1. Further ablation studies verify the effectiveness of the semantic
and appearance cues, their agreement mechanism, and
caption-conditioned detection. The main contributions of this work are summarized as follows.
\begin{itemize}
\item We formulate change-caption-guided RSCD using change
captions instead of change masks for training. 

\item We propose \method{}, a Semantic-Appearance Agreement Framework. It combines caption-grounded semantics with appearance changes for localization and conditions prediction on the same transition semantics.

    \item We develop a change-caption-driven bi-temporal image
    generation method that enables controllable source-to-target
    land-cover transitions, and use it to construct the Flair-RSGen
    dataset.

    \item We evaluate \method{} on Flair-RSGen and WHU-CDC. The results
    show higher macro-averaged IoU and F1 than the closest reproduced baselines, while the ablation studies verify
    the effectiveness of the proposed components.
\end{itemize}

\section{Related Work}
\subsection{Remote-Sensing Change Detection Under Limited Supervision}
Limited-supervision RSCD methods must recover spatial and semantic information that coarse labels do not provide. They then often rely on more complex models. For example, WSLCD combines image-level labels with self-supervised representations \cite{wslcd}. ACWCD and TransWCD refine activation maps with Transformer attention, change priors, and scene-adaptive constraints \cite{acwcd,transwcd}. BDFR-Net uses background consistency to separate and refine foreground responses, while BARNet adds teacher--student distillation and boundary-aware refinement \cite{bdfrnet,barnet}. SemSAM-CD converts activation peaks and boxes into segmentation prompts \cite{semsamcd}. AdvCP, SemiCD-VL, and SCM use adversarial prompts, vision--language priors, or target-label-free inference \cite{advcp,semicdvl,scm}.

These methods derive locations from image features or pretrained visual priors, and their image-level labels do not specify the semantic transition; existing refinement methods mainly improve the completeness or boundary quality of class activation maps, and cannot recover transition information that is absent from the supervision itself. In particular, a changed scene may contain several land-cover categories and several radiometric differences, while the binary label treats them alike. SAAF, in contrast, turns each ordered change caption into transition-specific spatial supervision, narrowing the task to one source-to-target event before spatial localization begins.
\begin{figure*}[t]
    \centering
    \includegraphics[width=\textwidth]{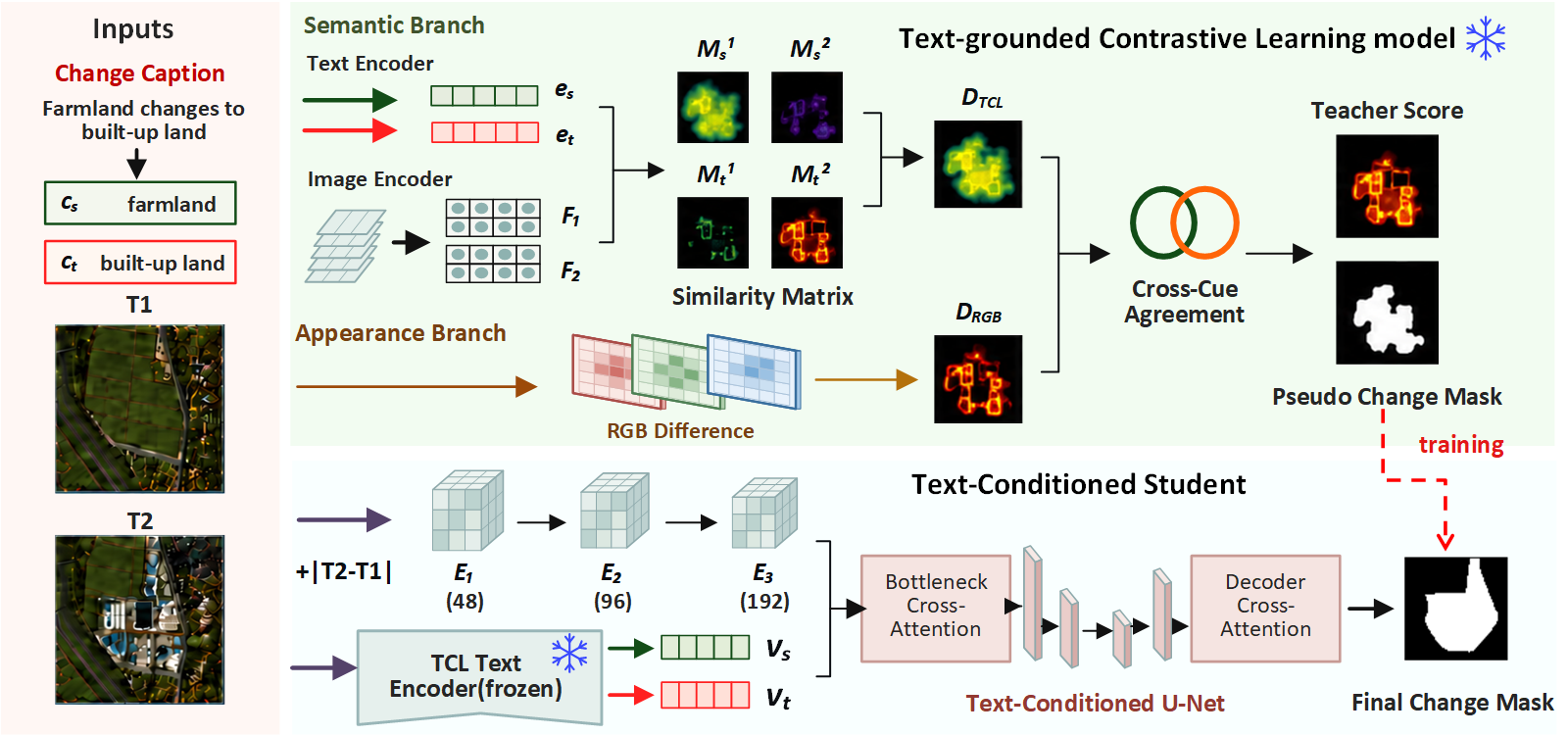}
    \caption{Overview of \method{}. Frozen TCL-DA grounds the caption concepts in both dates to form $\dtcl$, which is combined with $\drgb$ to produce score $S$ and pseudo mask $Y$. The student uses a single three-stage encoder with nine-channel input $[I_1;I_2;|I_2-I_1|]$. Its text branch projects frozen TCL-DA features into $v_s,v_t$. Their concatenation produces the additive bottleneck condition $h_c$, while tokens $T$ guide cross-attention at the bottleneck, after decoding, and within change refinement. Source and target heads provide auxiliary semantic supervision.}
    \label{fig:overview}
\end{figure*}

\subsection{Vision--Language and Caption-Aware Change Detection}
CLIP established scalable image--text representation learning \cite{clip}. RemoteCLIP and GeoRSCLIP adapt this principle to remote sensing \cite{remoteclip,georsclip}. SkyScript expands remote-sensing image--text pretraining data, while SkySense-O and CLIP2RS extend language-guided recognition to spatial interpretation \cite{skyscript,skysenseo,clip2rs}. These studies motivate domain-adapted text grounding but do not address caption-supervised bi-temporal change detection.

Language-guided prediction provides a link between image--text learning and spatial localization. GroupViT learns semantic grouping from image--text supervision~\cite{xu2022groupvit}, and MaskCLIP extracts responses from pretrained CLIP features~\cite{zhou2022maskclip}. LSeg aligns pixel features with text embeddings~\cite{li2022lseg}, DenseCLIP introduces pixel--text matching and contextual prompts~\cite{rao2022denseclip}, and CLIPSeg predicts regions specified by text or image prompts~\cite{luddecke2022clipseg}. Their supervision settings differ: language-guided inference does not itself imply training without labels. These studies also mainly concern concepts in a single image. RSCD requires evidence that the named source-to-target transition occurs across two dates.

Vision--language RSCD methods use text in several ways. ChangeCLIP predicts changes from a fixed land-cover vocabulary \cite{changeclip}; MDS-Net retrieves prompts from a predefined bank \cite{mdsnet}; and referring change detection localizes changes specified by a query \cite{referringcd}. Their predictors still learn from spatial labels. Although these methods incorporate textual semantics into change detection, text mainly serves as an auxiliary condition for feature learning, while their change predictors are still trained with real or generated change masks. They therefore do not address whether a
change detector can be trained from change captions alone.

\method{} uses a different supervision path. It receives an image pair and an ordered change caption, grounds the caption in both dates, verifies the response with appearance change, and trains the detector without reference masks.

\subsection{Synthetic Bi-Temporal Data Generation}
Synthetic data reduce the cost of collecting temporal pairs. Changen simulates semantic changes from a single image and semantic map, and Changen2 extends the process with a diffusion Transformer \cite{changen,changen2}. HySCDG combines real and generated imagery with semantic maps and change masks \cite{hyscdg}. WHU-GCD, RDF-MIG, and SNV-GenCD use vision--language, diffusion, or structural priors to generate paired images and spatial labels \cite{whugcd,rdfmig,snvgencd}. These labels are then used as detector supervision.

General image generation has also extended to image editing, where user guidance controls changes to an existing image. SDEdit uses a diffusion prior for guided editing~\cite{meng2022sdedit}. Blended Diffusion combines text guidance with a selected edit region~\cite{avrahami2022blended}, while Prompt-to-Prompt controls edits through cross-attention maps~\cite{hertz2023prompt}. These methods motivate local control, but do not by themselves define a remote-sensing transition dataset. Such a dataset requires the source class, target class, edit region, and retained caption to describe the same land-cover event.

We develop a text-guided image editing pipeline for remote sensing to construct Flair-RSGen. Each change caption specifies a source-to-target land-cover transition. The pipeline generates the corresponding image pair. It also produces corresponding change masks and semantic masks for evaluation. 

\begin{figure*}[t]
    \centering
    \includegraphics[width=\textwidth]{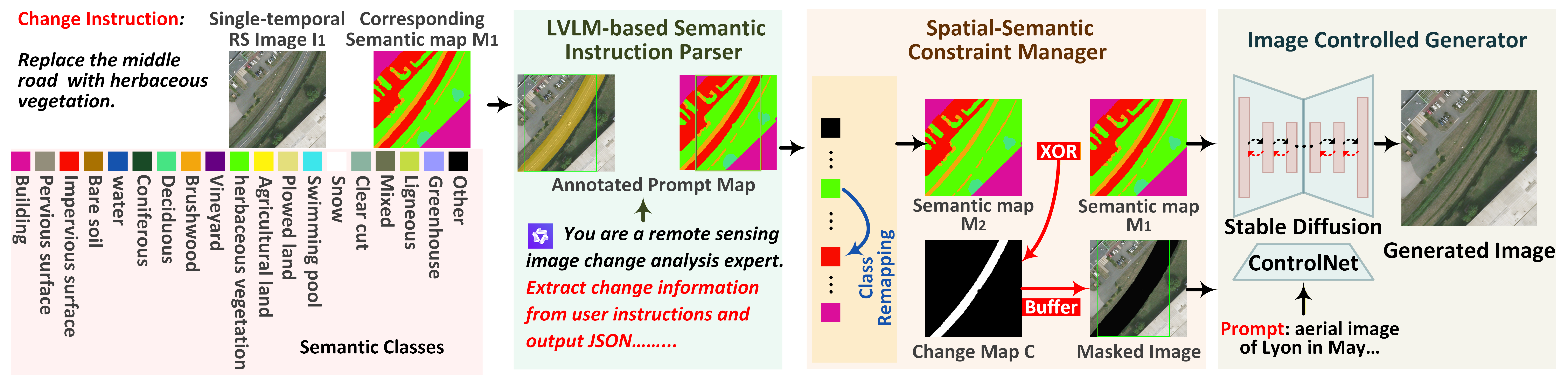}
    \caption{Flair-RSGen construction. A change instruction is parsed into semantic and spatial constraints, the selected source region is reassigned to the target category, and controlled generation modifies only the constrained neighborhood.}
    \label{fig:flair_pipeline}
\end{figure*}
\section{Proposed Method}
\subsection{Problem Formulation and Framework Overview}

Given two co-registered remote sensing images $I_1$ and $I_2$
together with a sample-specific change caption $c$, our objective is
to predict the region corresponding to the described land-cover
transition. A caption parser $\mathcal{P}$ decomposes $c$ into
\begin{equation}
(c_s,c_t)=\mathcal{P}(c),
\label{eq:caption_parser}
\end{equation}
where $c_s$ and $c_t$ denote the source concept before the change
and the target concept after the change, respectively. The parser also recognizes captions that explicitly indicate no
semantic change; no separate image-level label is required.

As shown in Fig.~\ref{fig:overview}, \method{} has a pseudo-mask teacher and a text-conditioned student. The teacher grounds $c_s$ and $c_t$ with a frozen, domain-adapted Text-grounded Contrastive Learning model (TCL-DA) to form $\dtcl$. In parallel, the appearance branch computes $\drgb$. Their agreement produces a teacher score $S$ and pseudo mask $Y$.

The two teacher branches are complementary. The semantic branch selects regions related to the named classes but can respond to unchanged instances. The appearance branch detects temporal variation but has no class information. Their agreement retains pixels supported by both cues and supplies the student with a caption-specific target.

The student concatenates $I_1$, $I_2$, and $|I_2-I_1|$ into a nine-channel tensor and processes it with a single three-stage visual encoder. Separate trainable projections map frozen TCL-DA text features to source and target vectors. These vectors form an additive bottleneck condition and ordered tokens for cross-attention. The tokens guide the bottleneck, decoder output, and coarse-to-refined change prediction. Two auxiliary heads predict source and target classes. This design carries the caption direction from pseudo-mask construction to final detection.

Pseudo masks are generated offline. During inference, the frozen teacher and pseudo-mask construction are not required. The student receives the image pair and its caption and outputs the final binary change map.

\subsection{Flair-RSGen Construction}
\subsubsection{Caption and Semantic Parsing}

Flair-RSGen begins with a FLAIR aerial image $I_1$ and its
semantic map $\mathbf{M}_1$ \cite{flair,flairmodel}.
A large vision–language model (LVLM) parser \cite{qwen} converts a change instruction into
\begin{equation}
(c_s,c_t,p)=\Phi(I_1,\mathbf{M}_1,c),
\label{eq:generation_parser}
\end{equation}
where $p$ is a spatial phrase, and $c_s$ and $c_t$ follow the
source--target convention in Eq.~(\ref{eq:caption_parser}). Synonyms
are normalized to the 19 FLAIR categories, while $p$ is grounded as
a coarse region $B(p)$. For example, given the instruction
``Agricultural land in the upper-right region changes to buildings,''
the parser assigns $c_s$ to agricultural land, $c_t$ to building,
and $p$ to the upper-right region. Accordingly, $B(p)$ restricts the
subsequent image editing to that region.

\subsubsection{Spatial--Semantic Constraint}
Let $u$ index the image lattice. We intersect $B(p)$ with pixels satisfying $\mathbf{M}_1(u)=c_s$ and retain the connected component that best matches $p$; the resulting edit region is $\Omega$. The target semantic map is
\begin{equation}
\mathbf{M}_2(u)=
\begin{cases}
c_t, & u\in\Omega,\\
\mathbf{M}_1(u), & \text{otherwise}.
\end{cases}
\label{eq:semantic_edit}
\end{equation}
The exact synthetic change mask is
\begin{equation}
C(u)=\mathbf{1}[\mathbf{M}_1(u)\neq\mathbf{M}_2(u)]
     \equiv \mathbf{M}_1\oplus\mathbf{M}_2,
\label{eq:synthetic_mask}
\end{equation}
where $\mathbf{1}[\cdot]$ is the indicator function and $\oplus$ denotes pixel-wise categorical inequality, i.e., categorical XOR. A dilated binary support $R=\operatorname{Dilate}(C)$ defines the editable neighborhood, and $I_{\mathrm{mask}}=I_1\odot(1-R)$ removes that neighborhood from the conditioning image. Here and below, $\odot$ denotes element-wise multiplication.

\subsubsection{Controlled Image Generation}
The image generator combines latent diffusion, ControlNet, and the pretrained HySCDG prior \cite{stable,controlnet,hyscdg}. It receives $\mathbf{M}_2$, $I_{\mathrm{mask}}$, the caption, and geographic metadata. If $\widetilde I_2$ is its raw output, unchanged pixels are preserved by
\begin{equation}
I_2=R\odot\widetilde I_2+(1-R)\odot I_1.
\label{eq:image_composite}
\end{equation}
The caption is the only user-provided change specification. The parser maps it to $(c_s,c_t,p)$, and the semantic map identifies pixels where the transition is valid. Changing class phrases changes the transition; changing $p$ changes its location.

For dataset construction, an LVLM receives $I_1$, $\mathbf{M}_1$, and a task prompt. It proposes three geographically plausible single-transition instructions with explicit spatial cues. Each proposal must name a visible source class, a different target class, and one spatial region. A candidate is retained only if its source class occurs in the requested region. Failed candidates are replaced by the next proposal, and the source image is skipped only when all three candidates fail. This screening avoids edits that contradict the source semantics.
\begin{figure*}[t]
    \centering
    \includegraphics[width=0.92\textwidth]{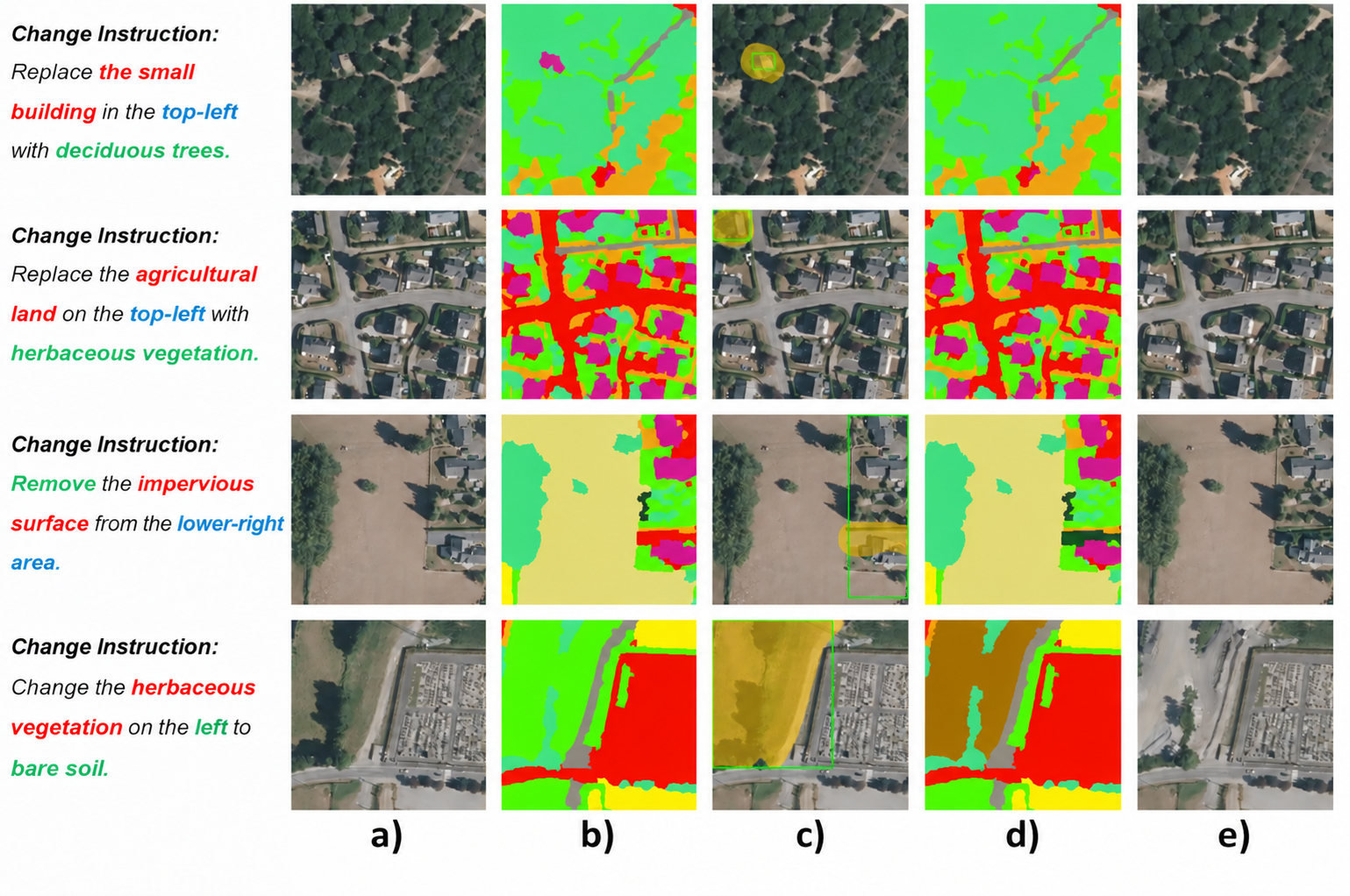}
    \caption{Flair-RSGen examples: instruction, first image, source map, selected region, target map, and generated second image.}
    \label{fig:flair_examples}
\end{figure*}

Real bi-temporal remote sensing images often exhibit appearance differences caused by illumination and seasonal variation. The caption-guided editing process above focuses on semantic transitions. To better reflect real acquisition conditions, we also use the same generator to construct no-change pairs. These hard negatives teach the detector to reject appearance differences without land-cover change. These pairs are generated separately from the
transition-based editing procedure in Eqs.~(3)--(5). These pairs provide negative examples
for learning to reject appearance differences
without a semantic transition. Fig.~\ref{fig:flair_pipeline} summarizes the pipeline, and Fig.~\ref{fig:flair_examples} shows representative samples.

\subsection{Semantic Branch: Directed Caption Evidence}
A change caption specifies the source and target concepts but does not provide their precise pixel locations. We use the frozen TCL-DA encoders to ground both concepts in each temporal image.

The prompts are $p_s=\text{``aerial image of }c_s\text{''}$ and $p_t=\text{``aerial image of }c_t\text{''}$. The frozen text encoder $E_T$ returns unit-normalized, 512-dimensional features:
\begin{equation}
e_q=E_T(p_q),\qquad q\in\{s,t\}.
\label{eq:text_embeddings}
\end{equation}
For each image $I_k$, TCL-DA extracts dense visual features
$F_k^{\mathrm{TCL}}$ for cross-modal matching. Their similarity
to the caption concept $q$ is
\begin{equation}
r_q^k(u)=
\frac{
F_k^{\mathrm{TCL}}(u)^{\top}e_q
}{
\lVert F_k^{\mathrm{TCL}}(u)\rVert_2
\lVert e_q\rVert_2+\epsilon
}.
\label{eq:tcl_similarity}
\end{equation}
TCL-DA converts these similarities into soft masks through
its pretrained mask-generation module. For $q\in\{s,t\}$
and $k\in\{1,2\}$, we define the final concept response as
\begin{equation}
M_q^k =
\mathcal{G}_{\mathrm{TCL\text{-}DA}}(I_k,e_q)
\in [0,1]^{H\times W},
\label{eq:tcl_response}
\end{equation}
where $\mathcal{G}_{\mathrm{TCL\text{-}DA}}$ denotes the frozen
soft-mask inference pipeline. It applies a learned affine
transformation and sigmoid activation to the similarity
scores, incorporates KP-branch fusion according to the
inference configuration, and resizes the output to the
input resolution. No image-wise min--max normalization
is applied to these individual response maps.

The four maps represent continuous support for the source
and target concepts at both dates. A source-to-target
transition should reduce support for the source concept
and increase support for the target concept. Following
this temporal pattern, we define the directed semantic
evidence as

\begin{align}
\dtcl=\mathcal{N}_{[0,1]}\!\Bigg(\frac{1}{4}\Big(&[M_s^1-M_s^2]_+
+[M_t^2-M_t^1]_+ \nonumber\\
&+|M_s^1-M_s^2|+|M_t^1-M_t^2|\Big)\Bigg),
\label{eq:tcl_change}
\end{align}
where $\mathcal{N}_{[0,1]}$ denotes image-wise min--max
normalization and $[a]_+=\max(a,0)$.

The first two terms model the expected source decrease and target increase. The absolute terms provide a symmetric fallback when the cross-modal response is reliable at only one date. Averaging the four terms reduces dependence on any single response map while preserving the ordered transition.

\subsection{Appearance Branch and Cross-Cue Agreement}
To complement the caption-guided transition evidence, we compute temporal RGB differences as a direct, class-agnostic cue of appearance change:
\begin{equation}
\drgb=\mathcal{N}_{[0,1]}\!\left(
\frac{1}{3}\sum_{r=1}^{3}|I_2^{(r)}-I_1^{(r)}|\right).
\label{eq:rgb_change}
\end{equation}
The raw agreement score is
\begin{equation}
S_{\mathrm{raw}} =
\sqrt{D_{\mathrm{TCL}}\odot D_{\mathrm{RGB}}}.
\label{eq:raw_agreement}
\end{equation}
We denote the processed score used for thresholding by $S$.
For Flair-RSGen, we apply a $5\times5$ average filter:
\begin{equation}
S = \operatorname{AvgPool}_{5\times5}(S_{\mathrm{raw}}),
\label{eq:score_smoothing}
\end{equation}
with the spatial resolution preserved. The pseudo mask
is then obtained as
\begin{equation}
Y(u)=\mathbf{1}[S(u)\geq\tau],
\qquad \tau=0.25.
\label{eq:pseudo_mask}
\end{equation}
No further min--max normalization is applied after smoothing.
For a no-change caption, $S$ and $Y$ are zero. For a no-change caption, $S$ and $Y$ are zero. The geometric mean suppresses semantic responses without appearance change and appearance differences unrelated to the caption.

The geometric mean favors locations supported by both semantic and appearance evidence. For example, a strong semantic response with little RGB variation receives a low score, as does a large RGB difference with weak support for the caption-defined transition. Arithmetic averaging can retain either response despite limited agreement. The geometric mean instead preserves jointly strong evidence, favoring changes consistent with both the caption and the observed image pair. We apply no further min--max normalization to avoid amplifying weak scores.

\subsection{Text-Conditioned Bi-Temporal Detector}
The detector concatenates the RGB images and their absolute difference along the channel dimension:
\begin{equation}
X=\operatorname{Concat}(I_1,I_2,|I_2-I_1|)\in\mathbb{R}^{9\times H\times W}.
\label{eq:detector_input}
\end{equation}
A single three-stage encoder processes $X$ to produce $E_1,E_2,E_3$, with 48, 96, and 192 channels. These joint temporal features are distinct from the teacher features $F_k^{\mathrm{TCL}}$.

The student reuses the frozen text features $e_s,e_t$ from Eq.~(\ref{eq:text_embeddings}). Independent trainable affine projections $P_s,P_t:\mathbb{R}^{512}\rightarrow\mathbb{R}^{64}$ give
\begin{equation}
v_s=P_s(e_s),\quad v_t=P_t(e_t),\quad v_c=[v_s;v_t].
\label{eq:student_text_projection}
\end{equation}
Here $v_s,v_t$ are the projected vectors shown in Fig.~\ref{fig:overview}, and $v_c\in\mathbb{R}^{128}$. Class indices select fixed TCL-DA features; no trainable class-embedding table is used. Ordered tokens and the bottleneck condition are
\begin{align}
T&=\operatorname{Stack}(v_s,v_t,v_t-v_s)\in\mathbb{R}^{3\times64},\nonumber\\
h_c&=W_2\delta(W_1v_c+b_1)+b_2\in\mathbb{R}^{192},\nonumber\\
\widetilde E_3&=\operatorname{CrossAttn}_{b}
\left(E_3+\operatorname{Broadcast}(h_c),T\right).
\label{eq:caption_conditioning}
\end{align}
The condition MLP has widths $128\rightarrow192\rightarrow192$, with ReLU $\delta$ between its affine layers. The condition is broadcast over the bottleneck's spatial dimensions before cross-attention. A decoder with skips from $E_2$ and $E_1$ produces $F_d$. A separate block then gives $\widetilde F_d=\operatorname{CrossAttn}_{d}(F_d,T)$.

Each cross-attention block uses visual features as queries and the three text tokens as keys and values. Its output is added to the visual input through a learned scalar gain. Bottleneck attention conditions low-resolution features, while decoder attention restores caption guidance after skip fusion.

The coarse head gives logits $Z_0=H_0(\widetilde F_d)$. The refinement head forms foreground- and background-weighted features from the detached coarse probabilities $\sigma(Z_0)$. It fuses these with $\widetilde F_d$, applies token cross-attention, and adds a learned residual to $Z_0$ to obtain $Z$. The prediction heads are
\begin{align}
P&=\sigma(Z),& Q_s&=H_s(\widetilde F_d),&
Q_t&=H_t(\widetilde F_d),\nonumber\\
\widehat{Y}&=\mathbf{1}[P\geq0.5].
\label{eq:prediction_heads}
\end{align}
Table~\ref{tab:network_architecture} summarizes the detector.

The refinement path is restricted to the change head. The source and target heads read the decoder feature directly and provide auxiliary class supervision on pseudo foreground pixels. The semantic heads operate alongside the change-refinement branch
and regularize the shared decoder features through auxiliary supervision.

\begin{table}[!t]
\centering
\caption{Architecture of the text-conditioned detector.}
\label{tab:network_architecture}
\scriptsize
\setlength{\tabcolsep}{3.0pt}
\renewcommand{\arraystretch}{1.12}
\begin{tabular}{
    >{\centering\arraybackslash}m{0.29\columnwidth}
    >{\centering\arraybackslash}m{0.22\columnwidth}
    >{\centering\arraybackslash}m{0.39\columnwidth}}
\toprule
Component & Input/Output & Configuration \\
\midrule
Input & $3+3+3\!\rightarrow\!9$ & $I_1,I_2,|I_2-I_1|$ \\
Single encoder & $9\!\rightarrow\!48\!\rightarrow\!96\!\rightarrow\!192$ & Joint temporal input; three two-block stages \\
Text projections & $512\!\rightarrow\!64$ each & $v_s=P_s(e_s)$, $v_t=P_t(e_t)$; frozen $e_s,e_t$ \\
Condition MLP & $128\!\rightarrow\!192\!\rightarrow\!192$ & $v_c=[v_s;v_t]$ to $h_c$; ReLU between affine layers \\
Bottleneck fusion & $192$ & Add broadcast $h_c$ to $E_3$; cross-attention over $T$ \\
Decoder & $192\!\rightarrow\!96\!\rightarrow\!48$ & Bilinear upsampling, skip fusion, and decoder cross-attention \\
Change heads & $48\!\rightarrow\!1$ & Coarse head; residual refinement with token cross-attention \\
Semantic heads & $48\!\rightarrow\!K$ each & Source and target logits \\
\bottomrule
\end{tabular}
\end{table}

\begin{table*}[!t]
\caption{Comparison on Flair-RSGen and WHU-CDC Test Sets Under the
Unified Image-Wise Macro-Averaging Protocol (\%)}
\label{tab:main_comparison}
\centering
\footnotesize
\setlength{\tabcolsep}{4.5pt}
\renewcommand{\arraystretch}{1.25}

\begin{tabular*}{\textwidth}{
@{\extracolsep{\fill}}lcccccccc@{}
}
\toprule
\rule[-1.1ex]{0pt}{3.6ex}
& \multicolumn{4}{c}{Flair-RSGen}
& \multicolumn{4}{c}{WHU-CDC} \\
\cline{2-5}\cline{6-9}
Method
& Macro IoU & Prec. & Rec. & F1
& Macro IoU & Prec. & Rec. & F1 \\
\midrule

WSLCD \cite{wslcd}
& 60.49
& \textbf{79.92}
& 64.06
& 67.09
& \underline{79.89}
& \textbf{84.35}
& 81.71
& 81.67 \\

ACWCD \cite{acwcd}
& \underline{64.52}
& 73.66
& \underline{73.40}
& \underline{70.46}
& 79.12
& 82.26
& \underline{86.08}
& \underline{82.22} \\

TransWCD \cite{transwcd}
& 39.15
& 39.68
& 72.25
& 42.72
& 24.63
& 30.43
& 28.22
& 27.60 \\

\method{} (Ours)
& \textbf{73.22}
& \underline{75.98}
& \textbf{92.90}
& \textbf{80.83}
& \textbf{81.55}
& \underline{82.33}
& \textbf{93.60}
& \textbf{84.28} \\

\bottomrule
\end{tabular*}
\end{table*}

\begin{table}[t]
\centering
\caption{Generation quality and semantic reliability of Flair-RSGen.}
\label{tab:generation_quality}
\setlength{\tabcolsep}{4pt}
\renewcommand{\arraystretch}{1.1}
\footnotesize
\begin{tabular}{lccccc}
\hline
Method
& FID $\downarrow$
& LPIPS$_{\mathrm{chg}}$ $\uparrow$
& LPIPS$_{\mathrm{bg}}$ $\downarrow$
& mIoU (\%) $\uparrow$
& PA (\%) $\uparrow$ \\
\hline
IP2P
& 158.83 & 0.516 & 0.528 & 21.7 & 41.3 \\
Flair-RSGen
& \textbf{8.21}
& \textbf{0.554}
& \textbf{0.138}
& \textbf{50.6}
& \textbf{71.7} \\
Real FLAIR
& Ref. & N/A & N/A & 59.9 & 78.3 \\
\hline
\end{tabular}
\end{table}

\subsection{Optimization Objective}
Let $\Omega_+=\{u:Y(u)=1\}$ and $\Omega_-=\{u:Y(u)=0\}$. For samples containing both sets, the mean foreground and background binary cross-entropies are
\begin{align}
\ell_+&=\frac{1}{|\Omega_+|}\sum_{u\in\Omega_+}\operatorname{BCE}(P(u),1),\\
\ell_-&=\frac{1}{|\Omega_-|}\sum_{u\in\Omega_-}\operatorname{BCE}(P(u),0).
\label{eq:positive_negative_bce}
\end{align}
The asymmetric map objective is
\begin{equation}
\lmap=0.35\ell_+ + 0.65\ell_-.
\label{eq:map_loss}
\end{equation}
If $\Omega_+$ is empty, $\lmap$ is the full-map background BCE; if $\Omega_-$ is empty, it is $\ell_+$. The larger background weight directly penalizes over-expanded predictions without suppressing the foreground through global class imbalance.

For nonempty $Y$, the overlap and semantic-alignment terms are
\begin{align}
\ldice&=1-\frac{2\sum_u P(u)Y(u)+\epsilon}
{\sum_u P(u)+\sum_u Y(u)+\epsilon},
\label{eq:dice_loss}\\
\lsem&=\frac{1}{2|\Omega_+|}\sum_{u\in\Omega_+}\big[
\operatorname{CE}(Q_s(u),c_s)+\operatorname{CE}(Q_t(u),c_t)\big],
\label{eq:semantic_loss}
\end{align}
where $\operatorname{CE}$ is multiclass cross-entropy. Both terms are set to zero for no-change samples. The final objective is
\begin{equation}
\mathcal{L}=\lmap+0.5\ldice+0.5\lsem.
\label{eq:full_loss}
\end{equation}
Here, $\ldice$ encourages region overlap, whereas $\lsem$ keeps the detected foreground consistent with the source-to-target transition. Their roles complement the spatial selectivity introduced by Eq.~(\ref{eq:raw_agreement}).

The region-normalized map loss prevents the much larger background from dominating optimization. Dice emphasizes the shape of nonempty pseudo masks, and the semantic term associates foreground features with the ordered classes. For no-change samples, only the background map term remains active, so these samples train rejection without inventing source or target pixels.

\section{Experiments}
\subsection{Datasets}
\subsubsection{Flair-RSGen}
Flair-RSGen is constructed from FLAIR aerial imagery and semantic
annotations \cite{flair,flairmodel} using
Eqs.~(\ref{eq:generation_parser})--(\ref{eq:image_composite}).
It contains both semantic-change pairs and no-change pairs.
Change pairs capture controlled transitions among 19 land-cover classes,
while no-change pairs preserve the semantic layout with appearance variation. Each pair includes a change caption, semantic maps for both dates, and a reference change mask. The dataset contains 45,761 pairs, split into 31,482 for training, 5,390 for validation, and 8,889 for testing. The dataset is publicly available at
\url{https://huggingface.co/datasets/qianyuancs/Flair-RSGen}.

\subsubsection{WHU-CDC}
WHU-CDC provides $256\times256$ bi-temporal image pairs, five captions per pair, image-level states, and binary masks \cite{whucdc}. We select one caption with an unambiguous source-to-target transition. The split contains 5,230 training, 650 validation, and 654 test pairs. Reference masks remain inaccessible to the training and validation loaders and are used only for final evaluation.

\subsection{Implementation Details}
Experiments use one NVIDIA RTX 4090 GPU with 24~GB memory. Images are resized to $256\times256$, and frozen TCL-DA generates semantic--appearance maps offline. The detector in Table~\ref{tab:network_architecture} uses one encoder on the nine-channel input, with stage widths of 48, 96, and 192. Frozen 512-dimensional TCL-DA features feed separate 64-dimensional projections. The same tokens condition the bottleneck, decoder, and change refinement. It is trained in FP32 for 30 epochs with AdamW, batch size 8, learning rate $5\times10^{-5}$, weight decay $10^{-4}$, and gradient clipping at 1.0. The checkpoint with the lowest pseudo-label validation loss is retained. The pseudo-mask and prediction thresholds are fixed by Eqs.~(\ref{eq:pseudo_mask}) and (\ref{eq:prediction_heads}).
\begin{figure}[!t]
    \centering
    \includegraphics[width=0.90\columnwidth]
    {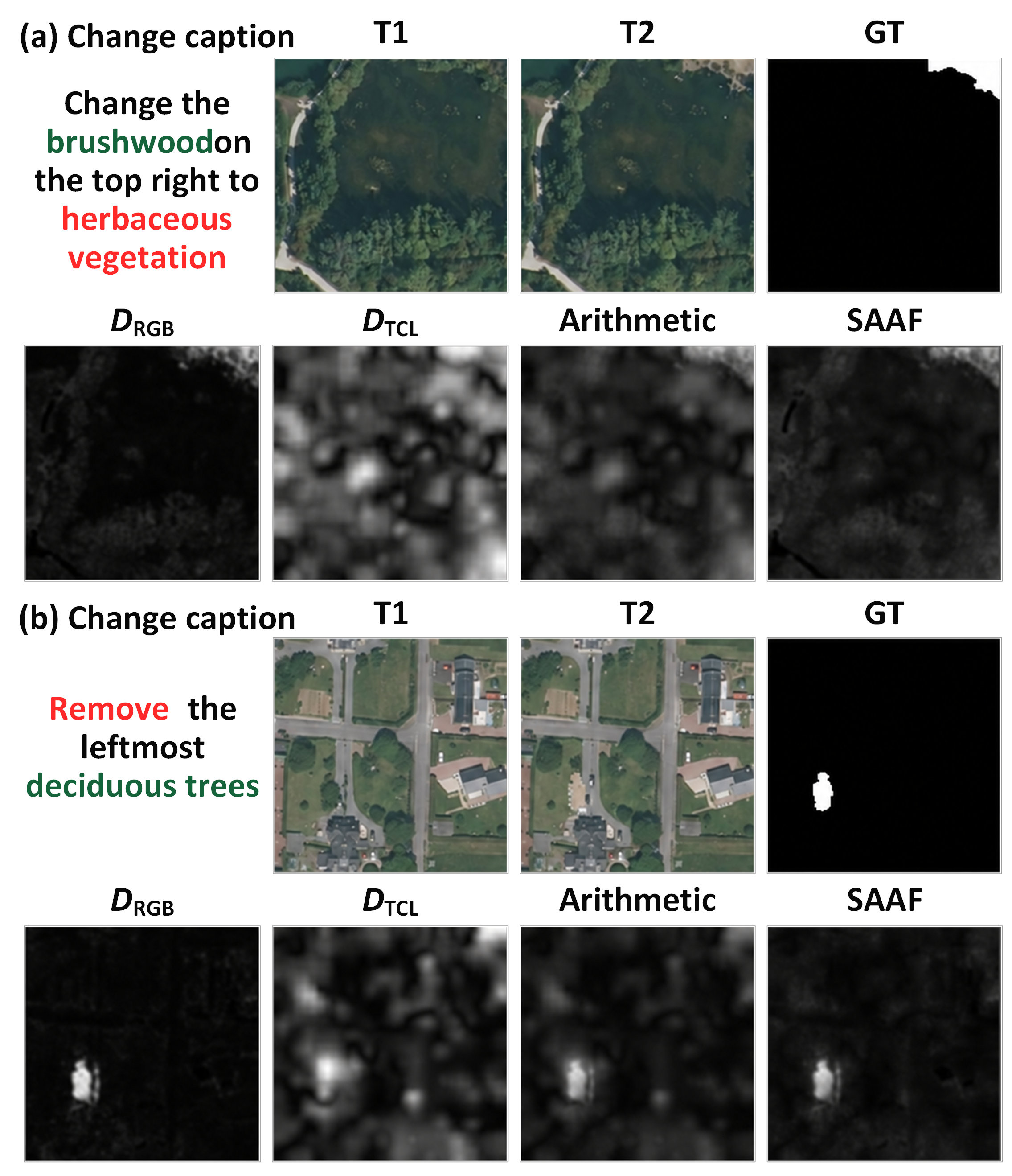}
    \caption{Semantic and appearance evidence. Brighter pixels indicate stronger change responses.}
    \label{fig:evidence_ablation}
\end{figure}
\begin{figure*}[!t]
    \centering
    \includegraphics[width=0.96\textwidth]{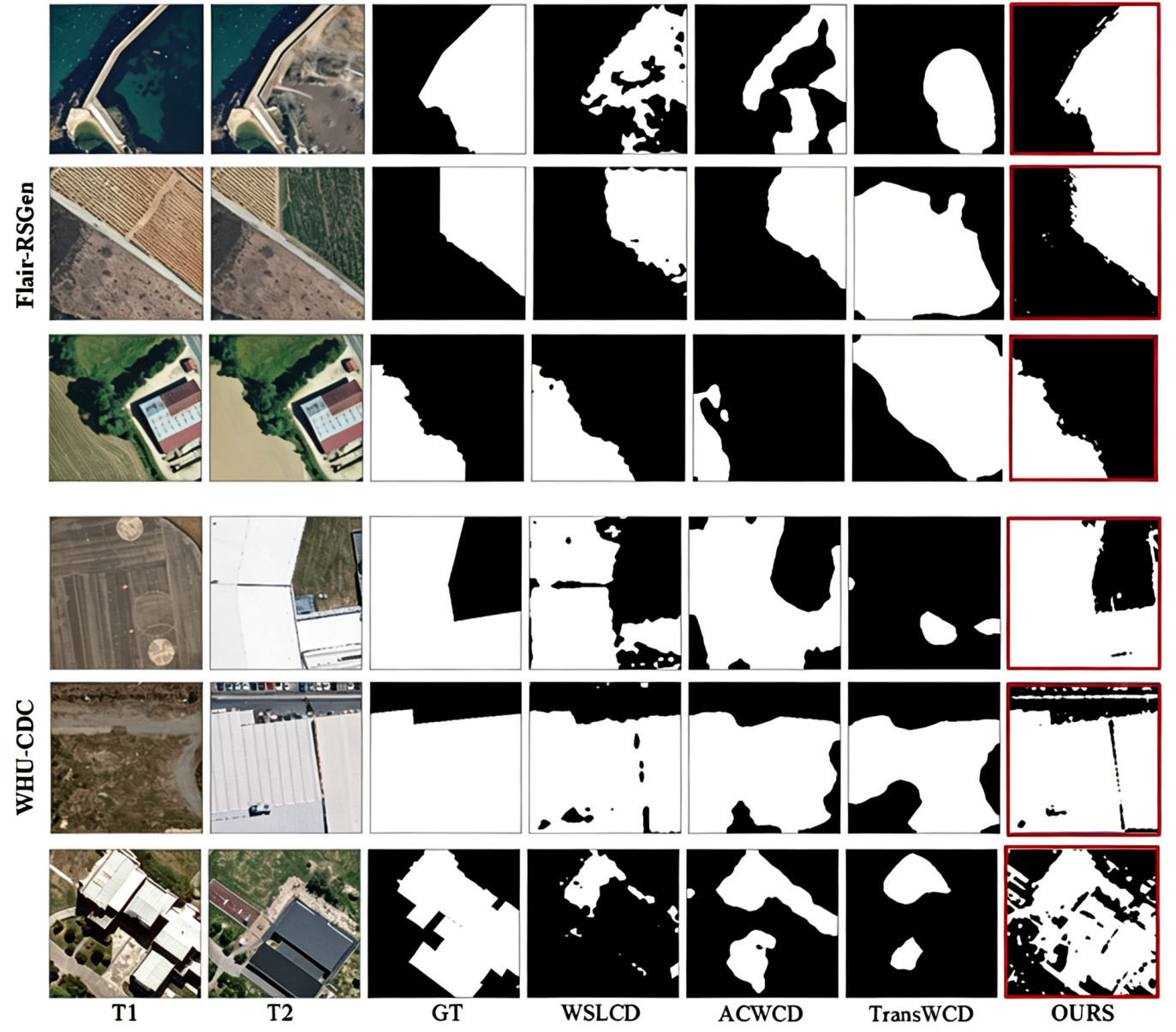}
    \caption{Qualitative results on Flair-RSGen (top) and WHU-CDC (bottom). White denotes changed regions.}
    \label{fig:qualitative_comparison}
\end{figure*}
TCL is adapted using a separate auxiliary set of approximately 5.8K generated Flair-RSGen samples. The adaptation set shares no source images with the detector training, validation, or test splits. Adaptation uses image–text supervision without semantic or change masks. The resulting TCL-DA image and text encoders remain frozen during detector training.

WSLCD, ACWCD, and TransWCD are retrained from their released implementations. All methods use the same image pairs, splits, and evaluation code. Each baseline retains its released backbone, optimizer, and inference rule. WSLCD uses ResNet-18 for 30 epochs. ACWCD and TransWCD use their released MiT-B1 configurations.

These baselines provide the closest reproducible limited-supervision comparison, although their supervision is not identical. They mainly learn from image-level change labels, whereas SAAF receives a sample-specific transition caption. We report Table~\ref{tab:main_comparison} under each method's native supervision. No reference mask enters optimization or checkpoint selection.

\subsection{Generation Quality and Reliability of Flair-RSGen}
We first test whether the generator produces realistic, localized edits. We use InstructPix2Pix (IP2P), a diffusion-based model that edits images following text instructions, as the baseline \cite{instructpix2pix}. Both methods are evaluated on 10,000 image pairs using
the same source images and editing instructions.
We measure distributional fidelity using the
Fr\'echet Inception Distance (FID)~\cite{heusel2017fid},
with real FLAIR imagery as the reference.
Learned Perceptual Image Patch Similarity
(LPIPS)~\cite{zhang2018lpips} is reported separately
for changed and unchanged regions to assess the extent
of editing and background preservation. Reading the measures together distinguishes realistic appearance, local control, and recognizable land-cover content.

Table~\ref{tab:generation_quality} shows that Flair-RSGen reduces FID from 158.83 to 8.21 compared with IP2P, indicating a generated image distribution closer to real FLAIR imagery. Changed-region LPIPS increases from 0.516 to 0.554, while background LPIPS decreases from 0.528 to 0.138. Together, these results indicate stronger edits within the intended region and better preservation of the surrounding scene. The fixed segmentation model achieves 50.6\% mIoU and 71.7\% pixel accuracy, compared with 21.7\% and 41.3\% for IP2P, suggesting that the generated land-cover classes are more recognizable. These scores also approach those on real FLAIR imagery (59.9\% mIoU and 78.3\% pixel accuracy), although a gap remains. Overall, the results support the visual realism, spatial control, and semantic consistency of Flair-RSGen.

\subsection{Evaluation Protocol}
We report image-wise macro IoU, precision, recall, and F1 on each test split. Metrics are computed per image and averaged. For a nonempty mask, binary definitions apply. For an empty reference mask, a metric equals one only if the prediction is empty; otherwise it equals zero. This rule evaluates localization and no-change rejection in the same score.

Image-wise averaging gives each pair equal weight and prevents large changed regions from dominating the score. We use the same empty-mask convention for every compared method. The protocol therefore measures both spatial overlap on changed pairs and complete rejection on no-change pairs.

\subsection{Comparison With Existing and Caption-Aware Baselines}
Table~\ref{tab:main_comparison} compares SAAF with limited-supervision RSCD methods under one evaluation protocol. On Flair-RSGen, SAAF obtains 73.22\% macro IoU and 80.83\% F1. It exceeds ACWCD, the strongest baseline on these metrics, by 8.70 and 10.37 points. The higher IoU indicates better overlap between
predicted and reference change regions. WSLCD achieves higher precision, but SAAF provides a better balance between missed changes and false detections.

On WHU-CDC, SAAF achieves the highest macro IoU, recall, and F1. It improves IoU by 1.66 points over WSLCD and F1 by 2.06 points over ACWCD. The gains are smaller than on Flair-RSGen, but show that SAAF also performs well on real bi-temporal images. These results demonstrate the value of caption-guided detection across both datasets. Fig.~\ref{fig:qualitative_comparison} provides a visual comparison of SAAF and the
baseline methods on both datasets. Since the methods use different supervision, we further examine caption conditioning through the matched comparisons below.

Because existing limited-supervision methods do not directly use sample-specific captions, Table~\ref{tab:matched_caption} provides a separate comparison with simple caption-aware baselines under matched supervision. Every row uses the same image pairs, captions, TCL-DA checkpoint, and data splits. Detector-based rows also share the training and evaluation protocol; the direct TCL-DA row does not train a detector. The plain detector retains the same architecture but uses
zero caption tokens and is trained without $\lsem$.

\begin{table}[!t]
\caption{Comparison With Supervision-Matched Caption-Aware Baselines on Flair-RSGen and WHU-CDC (\%)}
\label{tab:matched_caption}
\centering
\scriptsize
\setlength{\tabcolsep}{3.3pt}
\renewcommand{\arraystretch}{1.18}
\begin{tabular*}{\columnwidth}{@{\extracolsep{\fill}}lcccc@{}}
\toprule
& \multicolumn{2}{c}{Flair-RSGen} & \multicolumn{2}{c}{WHU-CDC} \\
\cline{2-3}\cline{4-5}
Variant & IoU & F1 & IoU & F1 \\
\midrule
TCL-DA direct & 61.08 & 68.21 & 79.78 & 82.22 \\
$\dtcl$ pseudo-mask + plain & 66.14 & 74.07 & 43.65 & 46.61 \\
Semantic only + conditioned & 66.77 & 74.52 & 80.58 & 83.14 \\
Agreement + plain detector & 61.83 & 69.40 & 17.27 & 20.22 \\
Agreement + conditioned ($\lsem=0$) & \underline{72.84} & \underline{80.48} & \textbf{81.59} & \textbf{84.31} \\
\method{} (full) & \textbf{73.22} & \textbf{80.83} & \underline{81.55} & \underline{84.28} \\
\bottomrule
\end{tabular*}
\end{table}

Table~\ref{tab:matched_caption}  examines whether captions are needed only for pseudo-mask generation or also for detector learning. We compare plain and text-conditioned detectors trained with the same pseudo labels. The plain detector receives no sample-specific caption information and uses no auxiliary semantic loss. The conditioned variant with $L_{\mathrm{sem}}=0$ allows us to examine text conditioning separately from this loss.

Using captions only to generate pseudo labels does not fully exploit their semantic information. With the agreement labels, adding text conditioning without $L_{\mathrm{sem}}$ raises IoU/F1 from 61.83/69.40 to 72.84/80.48 on Flair-RSGen, and from 17.27/20.22 to 81.59/84.31 on WHU-CDC. The pseudo labels and loss objective are unchanged
in these comparisons. The gains therefore show that the detector benefits from knowing which source-to-target transition it should detect, even when the training masks already come from captions.

Appearance agreement further improves results when the detector uses caption information. With the conditioned detector and training objective fixed, replacing semantic-only labels with agreement labels improves IoU/F1 by 6.45/6.31 percentage points on Flair-RSGen and 0.97/1.14 points on WHU-CDC. The same improvement is not observed for plain detectors. These results support using agreement supervision and text conditioning together: the teacher identifies regions supported by both cues, while the detector retains the transition semantics during prediction.

The auxiliary semantic loss has a relatively smaller effect. Adding $L_{\mathrm{sem}}$ improves Flair-RSGen IoU/F1 from 72.84/80.48 to 73.22/80.83, while WHU-CDC remains nearly unchanged at 81.55/84.28. Thus, the main benefit comes from text conditioning; the additional gain from semantic supervision depends on the dataset.

\FloatBarrier
\subsection{Ablation Studies}
Unless stated otherwise, ablations use Flair-RSGen with the same detector, training schedule, and evaluation protocol.
\begin{table}[!t]
\caption{Semantic--Appearance, TCL, and Objective Ablations on
Flair-RSGen Under the Image-Wise Macro Protocol (\%)}
\label{tab:ablation}
\centering
\footnotesize
\setlength{\tabcolsep}{3.8pt}
\renewcommand{\arraystretch}{1.22}

\begin{tabular*}{\columnwidth}{@{\extracolsep{\fill}}ccccc@{}}
\toprule
\rule[-1.0ex]{0pt}{3.4ex}Variant
& Macro IoU & Prec. & Rec. & F1 \\
\midrule

\multicolumn{5}{l}{
\rule[-0.8ex]{0pt}{3.6ex}\emph{Evidence ablation}} \\
Appearance only
& 68.74 & 77.86 & 79.55 & 76.25 \\
Arithmetic agreement
& 70.66 & 71.24 & \textbf{98.13} & 78.43 \\

\midrule
\multicolumn{5}{l}{
\rule[-0.8ex]{0pt}{3.6ex}\emph{TCL backbone ablation}} \\
Original TCL
& 71.84 & 76.15 & 88.75 & 79.20 \\

\midrule
\multicolumn{5}{l}{
\rule[-0.8ex]{0pt}{3.6ex}\emph{Complete framework}} \\
\method{} (full)
& \textbf{73.22} & 75.98 & 92.90 & \textbf{80.83} \\

\midrule
\multicolumn{5}{l}{
\rule[-0.8ex]{0pt}{3.6ex}\emph{Objective ablation}} \\
$\lmap+0.5\lsem$
& 71.79 & 73.66 & 94.75 & 79.49 \\
$\lmap+0.5\ldice$
& 72.84 & 75.44 & 93.18 & 80.48 \\
$\lmap$
& 71.30 & 72.98 & 95.00 & 78.91 \\

\bottomrule
\end{tabular*}

\end{table}
\begin{table}[!t]
\caption{Sensitivity to the Pseudo-Mask Threshold $\tau$ on
Flair-RSGen Under the Image-Wise Macro Protocol (\%).}
\label{tab:tau_sensitivity}
\centering
\footnotesize
\setlength{\tabcolsep}{4.0pt}
\renewcommand{\arraystretch}{1.18}
\begin{tabular*}{\columnwidth}{
@{\extracolsep{\fill}}ccccc@{}
}
\toprule
$\tau$ & Macro IoU & Prec. & Rec. & F1 \\
\midrule
0.15
& \textbf{73.39}
& \textbf{76.29}
& 92.57
& \textbf{81.01} \\
0.20
& 73.12
& 75.89
& 92.76
& 80.71 \\
0.25 (default)
& 73.22
& \underline{75.98}
& 92.90
& 80.83 \\
0.30
& 72.85
& 75.16
& \textbf{93.72}
& 80.48 \\
0.35
& \underline{73.34}
& 75.75
& \underline{93.66}
& \underline{80.94} \\
\bottomrule
\end{tabular*}
\end{table}
\subsubsection{Semantic and Appearance Evidence}
Table~\ref{tab:ablation} isolates the internal evidence-fusion choices. RGB difference alone misses caption-relevant changes. Arithmetic fusion reaches high recall but is expansive. Geometric agreement improves IoU and F1 over arithmetic fusion by 2.56 and 2.40 points because either branch can suppress an unsupported response.

Fig.~\ref{fig:evidence_ablation} illustrates how the two cues complement each other. Semantic responses highlight the named concepts but also cover nearby unchanged regions, while RGB differences help locate appearance changes without distinguishing their classes. Arithmetic fusion can carry these unsupported responses into the final map. By requiring agreement between the two cues, SAAF suppresses them and produces a more focused response around the vegetation edits shown in the figure. This visual comparison helps explain the IoU and F1 gains over arithmetic fusion reported in Table~\ref{tab:ablation}.

\subsubsection{TCL Domain Adaptation}
Replacing TCL-DA with the original TCL checkpoint while keeping Eqs.~(\ref{eq:tcl_change})--(\ref{eq:full_loss}) fixed reduces IoU by 1.38 points and F1 by 1.63 points. Although the original model gives higher precision, its lower recall shows that domain-adapted cross-modal responses cover aerial objects more completely.

\subsubsection{Objective Design}
On Flair-RSGen, the map loss alone gives the highest recall but lower IoU and F1, suggesting that it recovers most changes at the cost of more false positives. Adding semantic alignment improves IoU and F1 by 0.49 and 0.58 percentage points, indicating a modest benefit from preserving source--target semantics during training. Dice brings larger gains of 1.54 and 1.57 points, consistent with its role in improving overlap between predictions and pseudo masks. Combining both terms gives the best IoU and F1 on Flair-RSGen, showing that semantic alignment and overlap supervision provide complementary benefits. However, removing $L_{\mathrm{sem}}$ leaves WHU-CDC performance nearly unchanged in Table~\ref{tab:matched_caption}, so its benefit is not consistent across datasets.

\subsubsection{Threshold Sensitivity} Table~\ref{tab:tau_sensitivity} varies $\tau$ from 0.15 to 0.35. Macro IoU and F1 vary by only 0.54 and 0.53 percentage points, respectively, indicating that SAAF is stable across this threshold range. Although $\tau=0.15$ gives the best result, it improves over the default $\tau=0.25$ by only 0.17 IoU and 0.18 F1 points. The small differences suggest that the method does not require precise threshold tuning. We therefore retain $\tau=0.25$ as the common setting.

\section{Discussion}
The experiments reveal two complementary stages in caption-guided change localization. TCL-DA identifies the named transition, whereas RGB difference supplies direct evidence of temporal variation. Geometric agreement retains responses supported by both cues. Unlike arithmetic fusion, it prevents a strong response in one branch from compensating for weak evidence in the other.

Caption semantics remain useful after pseudo-mask construction. The matched controls show that agreement labels are most effective when the decoder receives the ordered source, target, and difference tokens. Cross-attention preserves the requested transition during spatial refinement, while $\lsem$ provides light, dataset-dependent regularization. This coupling accounts for the strong caption-conditioned agreement results on both datasets.

Flair-RSGen contributes controllable transitions and no-change hard negatives. Its distributional and semantic checks, together with the independent WHU-CDC evaluation, show that the framework can learn from generated pairs.

The framework also suggests clear extensions. Uncertainty-aware refinement could sharpen boundaries under registration offsets or gradual transitions. Multi-transition parsing could support captions containing several events, and broader geographic evaluation could further characterize transfer.

\section{Conclusion}
We presented \method{}, a semantic--appearance framework that turns change captions into supervision for remote sensing change detection, eliminating the need for manually annotated change masks during detector training. Frozen TCL-DA and RGB variation jointly construct spatial supervision, and the detector reuses ordered transition semantics through cross-attention. We also developed a caption-driven image generation pipeline to build Flair-RSGen at scale, producing controlled land-cover transitions and realistic no-change variations with paired captions, semantic maps, and change masks. The experiments demonstrate effective caption-guided learning on
Flair-RSGen and independently evaluate SAAF on observed
bi-temporal imagery from WHU-CDC. The matched controls further confirm that agreement-based pseudo supervision and caption-conditioned decoding form a complementary supervision path, with $\lsem$ acting as a light regularizer. Future work will extend the framework to boundary uncertainty, multiple transitions, and broader geographic settings.

\FloatBarrier
\balance
\bibliographystyle{IEEEtran}
\bibliography{references}

\end{document}